\documentclass[journal]{IEEEtran}

\usepackage{authblk}
\usepackage{graphicx}
\usepackage{svg}
\usepackage{hyperref}
\usepackage{graphicx}
\usepackage{caption}
\usepackage{subcaption}
\usepackage{lipsum}

\begin{document}

\title{Soft Labels for Rapid Satellite Object Detection}

\author[1]{\small Matthew Ciolino}
\author[1]{\small Grant Rosario}
\author[1]{\small David Noever}
\affil[1]{\footnotesize PeopleTec, Inc, 4901 Corporate Dr NW, Huntsville, AL 35805, USA}

\markboth{}%
{Shell \MakeLowercase{\textit{et al.}}: Bare Demo of IEEEtran.cls for IEEE Journals}

\maketitle

\begin{abstract}
Soft labels in image classification are vector representations of an image's true classification. In this paper, we investigate soft labels in the context of satellite object detection. We propose using detections as the basis for a new dataset of soft labels. Much of the effort in creating a high-quality model is gathering and annotating the training data. If we could use a model to generate a dataset for us, we could not only rapidly create datasets, but also supplement existing open-source datasets. Using a subset of the xView dataset, we train a YOLOv5 model to detect cars, planes, and ships. We then use that model to generate soft labels for the second training set which we then train and compare to the original model. We show that soft labels can be used to train a model that is almost as accurate as a model trained on the original data.
\end{abstract}

\begin{IEEEkeywords}
Soft Labels, Object Detection, Datasets
\end{IEEEkeywords}

\IEEEpeerreviewmaketitle

\section{Introduction}
Learning representations is a powerful tool in artificial intelligence. Deep learning has always been used to learn representations of images, text, and audio but furthermore can be used for transfer learning \cite{zhuang2020comprehensive} and pre-training \cite{mao2020survey}. In this way one models strength can be used to improve performance on another task. As is commonplace for many object detection backbones, a pre-trained feature extraction network \cite{he2015deep} is used to initialize the backbone of a model. This stabilizes training, improves convergence speed, and improves performance \cite{chen2019detnas}. Soft labels attempt this by abstracting the transfer of information to the dataset level.

Soft labels use a well-trained model to completely generate the training data for another model. A clear use of soft labels is in model distillation \cite{chen2017learning} where the final layer in a large neural network containing the class probabilities is used as the ground truth to train a smaller network on. This effectively teaches a smaller model what a large model learned in a teacher-student \cite{nguyen2022improving} relationship. In this paper, we use soft labels to train an object detection model on satellite imagery and then rapidly create a dataset of soft labels.

\subsection{Background}
Dataset creation can be a time-consuming and costly endeavor \cite{incze_2019}. While a handful of high-quality satellite object detection datasets exist \cite{chrieke}, they may not be sufficient for the task at hand. By automating the process of annotations through soft labels, we can cut down on the time it takes to make a dataset. Various problems and solutions arise with soft labels and here are some examples:

\begin{figure*}[t]
\centering
    \begin{subfigure}[h]{0.485\textwidth}
        \includegraphics[width=\textwidth]{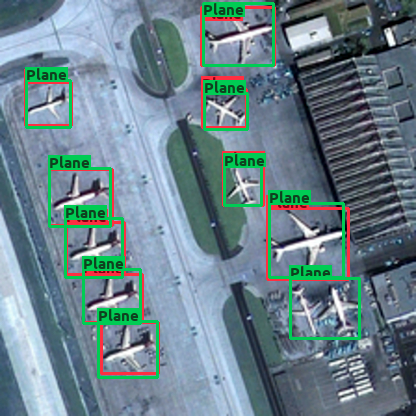}
        \caption{Train Set 1}
        \label{fig:fig1}
    \end{subfigure}
        \begin{subfigure}[h]{0.485\textwidth}
        \includegraphics[width=\textwidth]{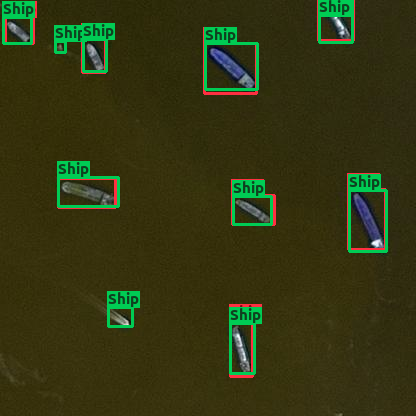}
        \caption{Train Set 2}
        \label{fig:fig2}
    \end{subfigure}
\caption{Ground Truth Labels (green) vs Soft Labels (red)}
\label{fig:soft_label_example}
\end{figure*}

\subsubsection{Missing Objects}
One intuitive problem with soft labels for object detection is that low confidence detections will be filtered not allowing a model to learn finer grain details for objects. While this would leave the objects in question to be designated as background objects, Wu et al. \cite{wu2018soft}, show that with the right modification to the bounding box loss function, a model can improve instead of worsening. To prove this, they dropped 30\% of the ground truth labels and found a 5\% drop in performance while when they weigh high-quality bounding boxes higher an increase of 3\% is found as compared to a baseline.

\subsubsection{Dataset Creation}
One effective use of soft labels for object detection is subtyping. Subtyping is the process of taking a class and breaking it down into smaller classes. For example, a car can be broken down into a sedan, truck, and SUV. In our past work, Rosario et al. \cite{rosario2022soft} show that with a simple car detection model and classifying cars by color, we can create a dataset of soft labels for car colors.

\subsubsection{Overfitting}
Overfitting is a common problem in deep learning. While there are many ways to combat overfitting, Zhang et al. \cite{zhang2021delving} proposed using an online label smoothing to generate soft labels more reliably. In each epoch during training, they mix hard labels and the previous epochs soft labels to iteratively improve the soft labels of each object detected. That mixture is governed by the cross-entropy classification loss with the original distribution of soft labels being uniform. This method is shown to better define and separate classes on image datasets bringing a 2.1\% gain on performance for top-1 error on VOC \cite{journals/ijcv/EveringhamGWWZ10} and COCO.

\subsubsection{Student/Teacher Models}
Using the soft labels from a teacher model, a student model can outperform models trained on partially labeled COCO data. Xu et al. \cite{xu2021end} in this semi-supervised object detection (SSOD) paper, presented 2 techniques: soft teacher, where the teacher model is actually updated by the student using an exponential moving average, and a training strategy where the student/teacher models are trained using images with different augmentations. This use of soft labels improves state-of-the-art performance by 8.43\% on 1\% to 10\% labeled coco data.

\subsection{Contributions}
The above methods are all examples of how soft labels can be used to improve a model. In this paper, we approach a far simpler task of simply investigating the performance drop-off using 100\% soft labels as compared to the complete dataset. Ground truth versus soft label dataset can be seen in [Figure \ref{fig:soft_label_example}]. In this paper we attempt to answer 3 research questions that pertained to the nuances of soft labels:

\begin{itemize}
\item How does the soft labeling affect performance?
\item Can this be used to categorize datasets automatically?
\item What confidence value trains the best soft label model?
\end{itemize}

\begin{figure}[t]
    \centering
    \includegraphics[width=.485\textwidth]{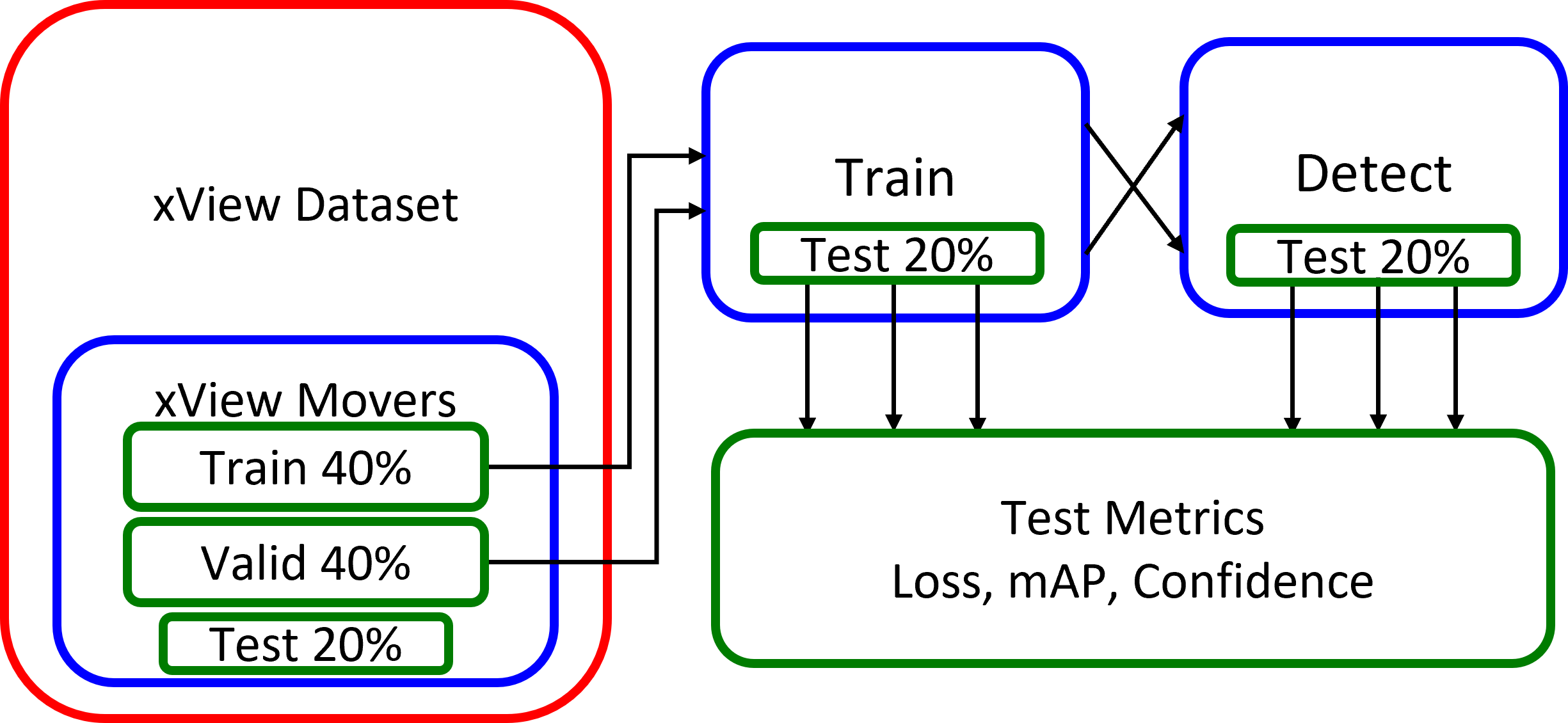}
    \caption{Expierment Flow}
    \label{fig:exp}
\end{figure}

\begin{table*}[ht]
\caption{Test Set Metrics}
\label{table1}
\centering
\begin{tabular}{||l|c|c|c|c|c|c|c||}
\hline
\multicolumn{1}{||c|}{\textbf{Model}} & \textbf{Conf} & \textbf{mAP50} & \textbf{mAP95} & \textbf{F1 Score} & \textbf{box\_loss} & \textbf{obj\_loss} & \textbf{class\_loss} \\ \hline
Train Set 1 & - & 0.76251 & 0.45069 & .71   @ .419c & 0.052841 & 0.03327 & 0.0058541 \\ \hline
Train Set 1 Soft & 0.3 & 0.73275 & 0.42357 & .70 @ .482c & 0.042647 (-21.35\%) & 0.029469 (-12.11\%) & 0.0033020  (-55.74\%) \\ \hline
Train Set 1 Soft & 0.5 & 0.70798 & 0.40687 & .69   @ .318c & 0.044096 (-18.04\%) & 0.030910   (-7.35\%) & 0.0041233  (-34.69\%) \\ \hline
Train Set 2 & - & 0.77470 & 0.46090 & .72 @ .421c & 0.0415 & 0.027994 & 0.002451 \\ \hline
Train Set 2 Soft & 0.3 & 0.72932 & 0.42006 & .70   @ .503c & 0.042767  (+3.01\%) & 0.031548 (+11.93\%) & 0.0033173 (+30.04\%) \\ \hline
Train Set 2 Soft & 0.5 & 0.71681 & 0.41578 & .70 @ .308c & 0.044652  (+7.32\%) & 0.030846  (+9.69\%) & 0.0044479 (+57.89\%) \\ \hline
\end{tabular}
\end{table*}


\begin{figure*}[t]
    \centering
    \includegraphics[width=.985\textwidth]{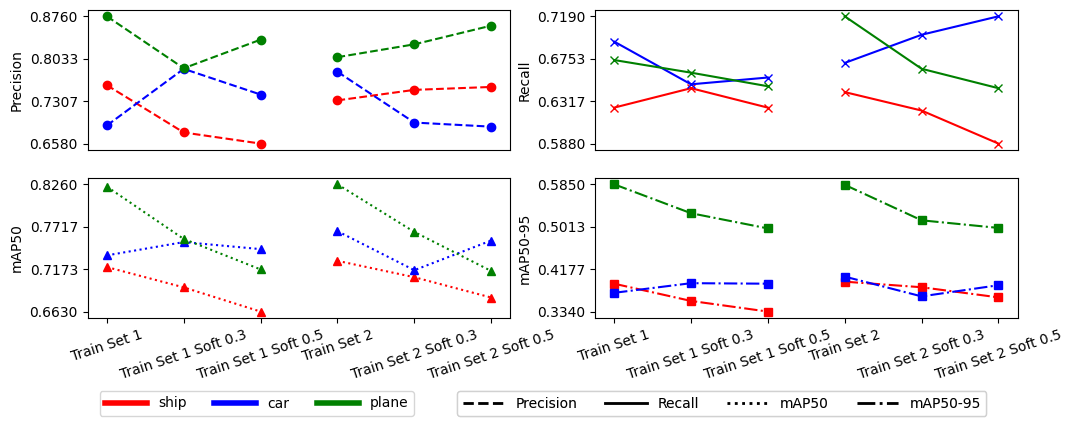}
    \caption{Per Class Metrics Plot - Color/Linetype to Class/Metric}
    \label{fig:per}
\end{figure*}

\section{Experiment}
To answer these questions, we lay out the following experiments [Figure \ref{fig:exp}] involving a subset of the xView dataset \cite{lam2018xview}. We sub-select xView for moving objects (vehicles, planes, and ships) leaving 34,252 images collected from parking lots, marine ports, and airports. Approximately one-third of the pictures were background, meaning they contained no examples but contributed to the training "null" set. Our data contains 37,712 instances of ships, 174,779 instances of cars, and 18,052 instances of planes. These counts are randomly split across 3 sets for train\_1/train\_2/valid sets containing 40/40/20 percent of the data respectively. More dataset information can be found in the appendix.

YOLOv5s \cite{jocher2022ultralytics} containing 7,027,720 parameters was trained on the train\_1/train\_2 set for 3 epochs with a batch size of 16 using the pretrained coco \cite{lin2014microsoft} model as a checkpoint. Using those models, we soft-labeled the opposite training set to generate soft-labeled train\_2/train\_1. All models trained used the unseen valid set as the test set to compare metrics. We use the mAP, F1, and losses to evaluate the models.

\begin{itemize}
\item mAP is the mean of the average precision value for recall value over 0 to 1 for each class at a given IoU. \cite{hui_2019}
\item The IoU threshold is the intersection over union of the predicted bounding box and the ground truth bounding box. \cite{baeldung_2022}
\item F1 is the harmonic mean of precision and recall at different confidence thresholds. \cite{deepai_2019}
\end{itemize}

We also look at train/test loss. For loss, YOLOv5 uses a combination \cite{lihi_gur_arie_2022} of bounding box loss, objectness loss, and classification loss to train the model. 

\begin{itemize}
\item The bounding box loss is the mean squared error between the predicted bounding box and the ground truth bounding box.
\item The objectness loss measures the probability that an object exists in a proposed region of interest through binary cross entropy.
\item The classification loss is the cross-entropy loss between the predicted class and the ground truth class. 
\end{itemize}

A detailed summary of YOLOv5 can be found here \cite{ultralytics}. The test set which the results are all based on is broken down as follows: Across the 7,896 images and 58,165 instances in the test set, 8,710 instances (15\%) are ships, 45,025 instances (77\%) are cars and 4,430 instances (8\%) are planes.

In addition, during experimentation OWL-ViT \cite{minderer2022simple} was released. This Open-Vocabulary Object Detection model can be used to soft label. In the appendix section, we compare YOLOv5 models with the COCO 2017 validation set with ground truth, 0.1 confidence OWL-ViT inference, and 0.25 confidence OWL-ViT inference.

\section{Results}

While ground truth models performed the best [Table \ref{table1}], the soft label models consistently came within a 6\% difference for mAP and F1 score. Interestingly, the F1 score remains very consistent (.68 to .70) with the optimal confidence value acting inversely to the confidence threshold of the soft-labeled training set. When looking at losses of the test set after the final epoch of training, we see a more interesting story that might tell us why the metrics are lower. 

While the soft label trained model losses are both higher and lower than the ground truth model, we can derive insight from the relative differences. Both bounding box loss and objectness loss seem relatively consistent at an average difference of 11.35\%. This is in stark contrast to the classification loss which averages a difference of 44.59\%. Since classification loss has the highest change (still the lowest component of the loss) it can be said that the unlabeled objects in the soft label training sets account for the loss of performance. 

We can take a look at the per-class metrics [Figure \ref{fig:per}] to get a better look at what might be going on. As shown, the majority of the decrease in performance comes from soft-labeled planes. For planes trained on 0.5 confidence soft labels, all metrics dropped on average 8.49\%. Interestingly, cars, which are the smallest but most abundant object in the dataset performed only 0.65\% worse across all metrics than the ground truth models. Overall, per-class metrics dropped 4.44\% across all metrics. 

\section{Discussion}

While this performance drop is statistically significant in certain cases, i.e., planes, the benefits of soft-labels outweigh this loss in performance by providing additional data in a low-cost and efficient manner as well as potentially increasing model knowledge in the cases of transfer learning. Moreover, this loss in performance can usually be remedied by either balancing the dataset or increasing data for lagging labels, as evidenced by the statistically insignificant drop in car labeling performance.

\section{Conclusion}
Models can be trained exclusively on soft labels with a less than 6\% drop in mAP as compared to ground truth labels on the same dataset. Regardless of the confidence threshold used to create the soft labels, mAP/F1 scores remain within 1\% of each other. This suggests that the soft labels are not overfitting to the training data. These results validate that rapid object detection datasets can be created with soft labels and that soft labels can be used to train models with a high degree of accuracy.

\section*{Future Work}
In future work, we would like to use soft labels to improve the performance of the model on the test set by supplementing existing data instead of training solely on the soft labels. We would also like to use soft labels from one generalized model that could categorize any satellite image you are looking at into a dataset.

\section*{Acknowledgment}
The authors would like to thank the PeopleTec Technical Fellows program for encouragement and project assistance.

\bibliographystyle{./bibtex/IEEEtran}
\bibliography{./IEEEexample}

\section*{Appendix}

Dataset information from the training sets. Instance counts bounding boxes, bounding boxes, box center locations, and box width and height.

\begin{figure}[htb]
\centering
    \begin{subfigure}[h]{0.45\textwidth}
        \includegraphics[width=\textwidth]{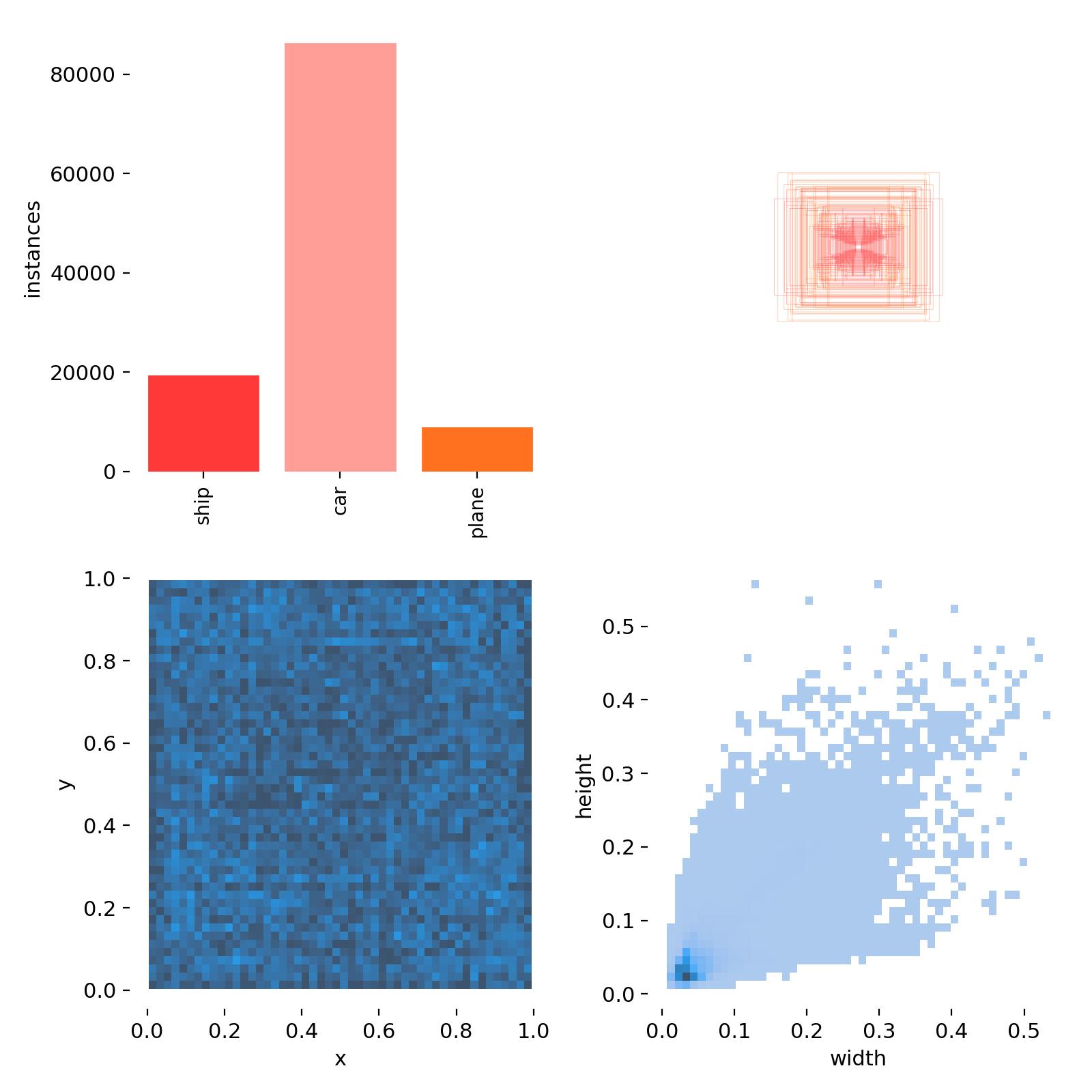}
        \caption{Train Set 1 Dataset Information}
        \label{figAp:fig1}
    \end{subfigure}
        \begin{subfigure}[h]{0.45\textwidth}
        \includegraphics[width=\textwidth]{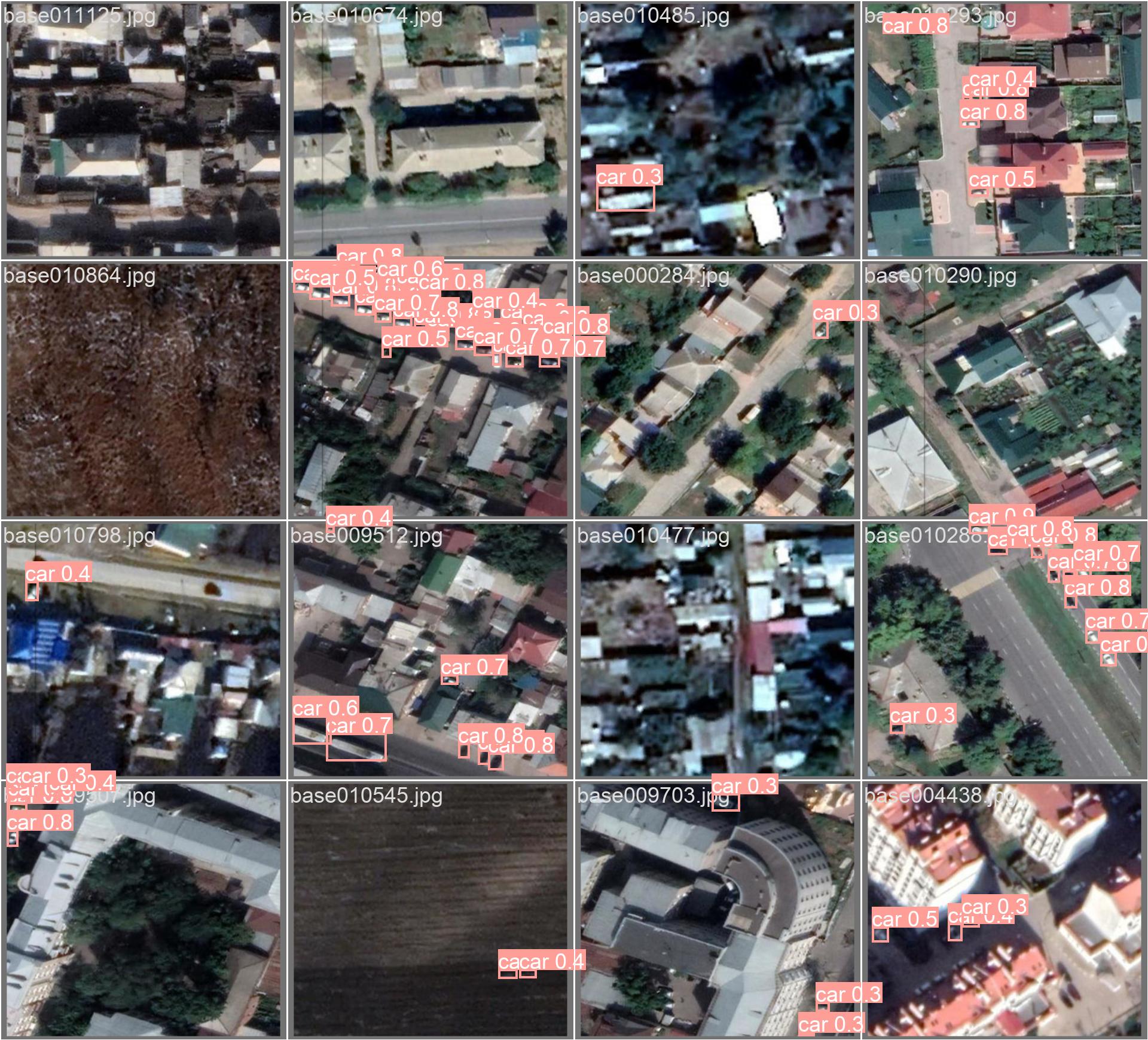}
        \caption{Train Set 1 Soft Labels}
        \label{figAp:fig2}
    \end{subfigure}
\caption{Dataset Information (a) and Example Predictions (b)}
\label{figAp:subfigureexample}
\end{figure}

\begin{figure*}[ht]
    \centering
    \includegraphics[width=0.85\textwidth]{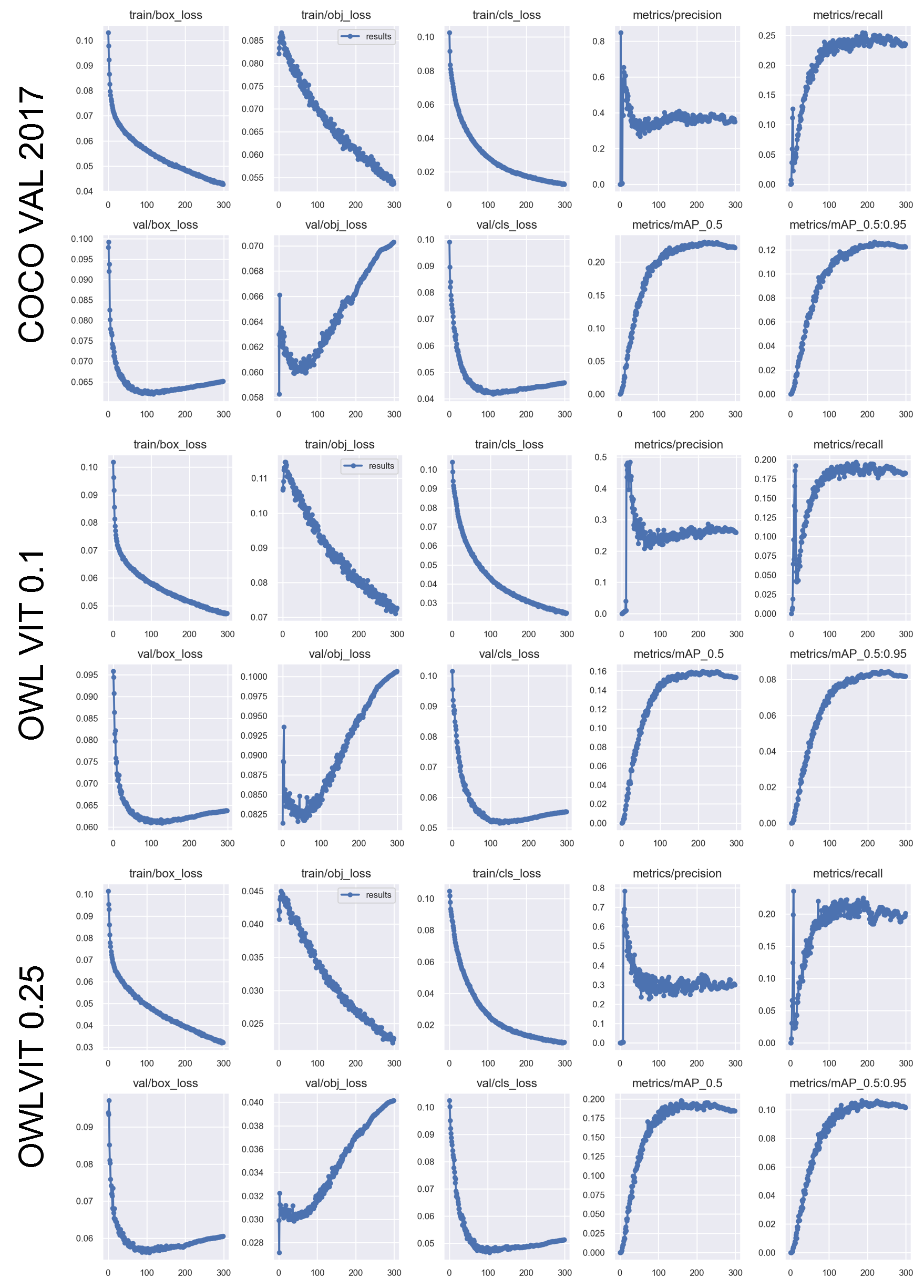}
    \caption{OWL-ViT YOLOv5 COCO-VAL GT, 0.1, 0.25 Training Metrics}
    \label{fig:owlvit}
    As shown, mAP at 0.5 IoU for the ground truth is 0.2225, OWL-ViT at 0.25 confidence is .1844 (-17.12\%), and OWL-ViT at 0.10 confidence is .1534 (-31.05\%). Results for this experiment can be found at this public tensorboard: \url{https://tensorboard.dev/experiment/J5prk5YeRKqZEuLPigU7jw/#scalars} 
\end{figure*}

\end{document}